\documentclass[letterpaper, 10 pt, conference]{ieeeconf}  

\IEEEoverridecommandlockouts                              

\usepackage{epsfig}
\usepackage{multirow}
\usepackage{makecell}
\usepackage{amsmath}
\usepackage{amssymb}
\usepackage{times}
\usepackage{pifont}
\newcommand{\cmark}{\ding{51}}
\newcommand{\xmark}{\ding{55}}
\usepackage[ruled,vlined]{algorithm2e}
\usepackage[bookmarks=false]{hyperref}
\usepackage{subcaption}

\usepackage{xcolor}
\usepackage{esvect}

\title{\LARGE \bf
Uncertainty-aware Self-supervised 3D Data Association
}

\author{Jianren Wang$^{1}$, Siddharth Ancha$^{1}$, Yi-Ting Chen$^{2}$ and David Held$^{1}$
\thanks{$^{1}$Jianren Wang, Siddharth Ancha, David Held are with the Robotics Institute, Carnegie Mellon University, 5000 Forbes Ave., Pittsburgh, PA 15213, USA
        {\tt\small jianrenw, sancha, dheld@andrew.cmu.edu}}%
\thanks{$^{2}$Yi-Ting Chen is with the Honda Research Institute, 375 Ravendale Dr, Mountain View, CA 94043, USA
        {\tt\small ychen@honda-ri.com}}%
}

\begin{document}

\maketitle
\thispagestyle{empty}
\pagestyle{empty}

\begin{abstract}
3D object trackers usually require training on large amounts of annotated data that is expensive and time-consuming to collect. Instead, we propose leveraging vast unlabeled datasets by self-supervised metric learning of 3D object trackers, with a focus on data association. Large scale annotations for unlabeled data are cheaply obtained by automatic object detection and association across frames. We show how these self-supervised annotations can be used in a principled manner to learn point-cloud embeddings that are effective for 3D tracking. We estimate and incorporate uncertainty in self-supervised tracking to learn more robust embeddings, without needing any labeled data. We design embeddings to differentiate objects across frames, and learn them using uncertainty-aware self-supervised training. Finally, we demonstrate their ability to perform accurate data association across frames, towards effective and accurate 3D tracking. Project videos and code are at \url{https://jianrenw.github.io/Self-Supervised-3D-Data-Association/}.

\end{abstract}

\section{Introduction}

3D object tracking is the problem of detecting and associating objects in multiple frames of a sequence of 3D point cloud data. For example, autonomous  vehicles must continually sense their environment via sensors such as LIDARs. These sensors generate  sequences of 3D data, from which the vehicle must estimate the locations of objects in the environment surrounding the vehicle. Further, in order to safely navigate in its surroundings, perception algorithms must be able to track dynamic objects, such as cars, vehicles and pedestrians.  3D tracking involves simultaneously detecting objects in the current frame, as well as associating them with objects seen in previous frames. 

State-of-the-art tracking algorithms employ convolutional neural networks for 3D tracking~\cite{luo2018fast,yang2018pixor}. However, a major problem with such approaches is that they require vast amounts of labeled data because they are trained using supervised learning. Obtaining large-scale, human-annotated labeled data, which includes bounding box annotations and associations across frames, can be expensive and time-intensive. On the contrary, vast amounts of \textit{unlabeled} data is cheaply available. For example, LIDAR sensors fitted on vehicles continuously collect and store data streams without requiring any human processing.

In this work, we propose to leverage unlabeled data via self-supervised learning of 3D trackers without the need for any labeled data. The idea is to first run a learned object detector on large unlabeled datasets in an automated fashion, as well as to automatically associate detections across frames. These labels are then used as ``pseudo-ground truth" in a supervised way to learn 3D embeddings of objects. Since detection and association is automated, it is trivial to obtain large quantities of pseudo-ground truth labels on unlabeled datasets.  This data can then be used to train deep neural network embeddings suitable for data association for 3D tracking.

\begin{figure}
    \centering
    \includegraphics[width=\linewidth]{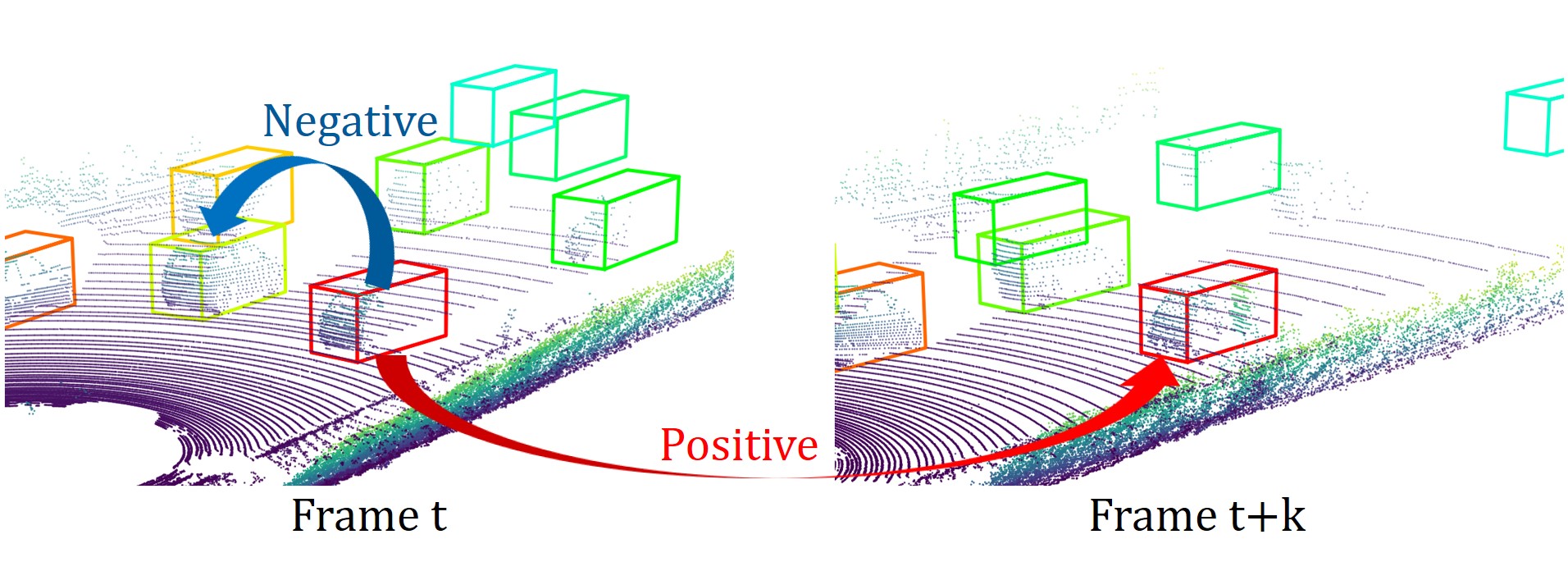}
    \caption{Example of a triplet during self-supervised training. We use a fully automatic pipeline to generate training pairs and learn in an uncertainty-aware manner from unlabeled data.}
    \label{fig:pull}
\end{figure}

We demonstrate a technique to learn 3D embeddings of point clouds of objects that capture the 3D geometric properties unique to that object instance. These embeddings are designed to be similar for detections of the same object observed at different time instants (different frames) but dissimilar for detections of different objects. The embeddings can then be used to perform associations of bounding boxes for 3D tracking -- detections with similar embeddings can be associated with each other as they likely come from the same object. We show that self-supervised learning on unlabeled data can learn 3D embeddings that are effective for associating bounding boxes across frames while performing 3D tracking.


A potential limitation with learning using labels generated by automated techniques is that these labels are not necessarily reliable. The detection system or the automated association algorithm may produce mistakes, and learning on these mistakes may lead to inferior 3D embeddings. We propose to explicitly tackle this problem by incorporating \textit{uncertainty} into self-supervised learning. Specifically, we propose a method for estimating how uncertain the self-supervised association algorithm is during training. We then weight the supervised loss using this uncertainty -- if the uncertainty on a training example is high, its contribution to the loss will be small, and vice versa. This helps avoid learning a poor embedding due to incorrectly labeled examples. We show that incorporating uncertainty estimates during self-supervised learning helps learn more robust 3D embeddings that produce superior tracking performance.

We summarize our contributions as follows:
\begin{itemize}
    \item We propose using self-supervised learning on large unlabeled datasets to learn 3D embeddings for object tracking, without any labeled data. 
    \item We show that self-supervised 3D embeddings can effectively be used to associate objects across frames
    \item We show that it is possible to leverage \textit{uncertainty estimates} during self-supervised tracking to learn more robust  3D embeddings.
    
\end{itemize}



    
    
    
    
    
    

\begin{figure*}
    \centering
    \includegraphics[trim=0 0 0 0,clip,width=0.98\textwidth]{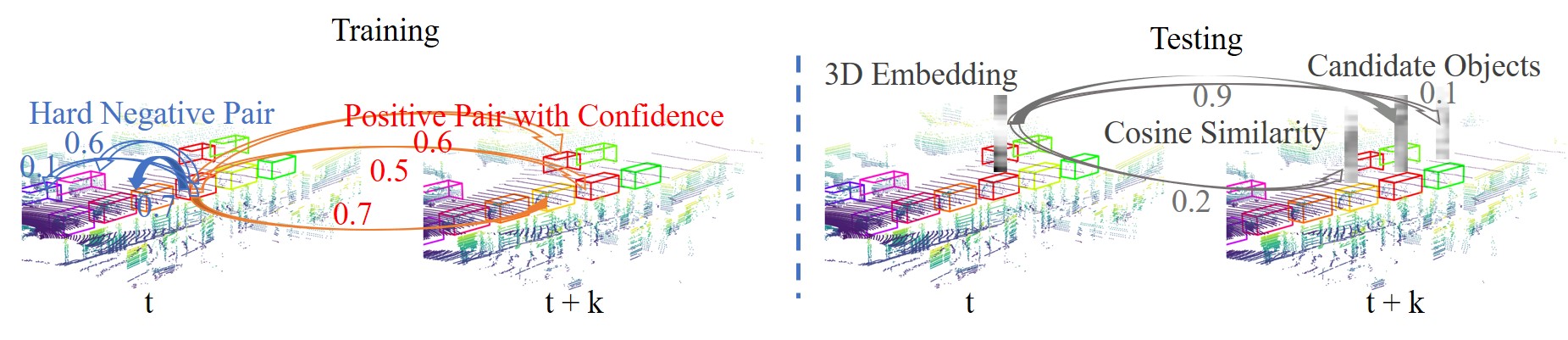}
    \caption{\textit{Left:}
    Triplet example during self-supervised training. For any anchor detection in frame $t$, we select the hardest negative example from the same frame whose embedding produces the largest cosine similarity with the anchor detection. A positive example is picked from a detection in another frame that is associated with the same track as the anchor detection. A confidence of association is estimated and used to weight this example during self-supervised training. We train the embedding network to maximize the agreement between associated pairs. \textit{Right}: At test time, self-supervised embeddings are extracted from each candidate detection in a frame. We use cosine similarity of embeddings extracted from each pair of objects to represent their appearance similarity, which is further used to perform accurate data association across frames.}
    \label{fig:pipeline}
\end{figure*}

\section{Related Work}

\subsection{Self-Supervised Learning}
One form of self-supervised learning deals with predicting a subset of data from another subset. 
Prior works perform such self-supervised learning via applying transformations for data augmentation~\cite{dosovitskiy2015discriminative}, predicting transformations that were applied such as rotation~\cite{gidaris2018unsupervised}, predicting relationship between patches of an image~\cite{noroozi2016unsupervised,noroozi2017representation} or frames of a video~\cite{fernando2017self,wei2018learning}, colorization~\cite{zhang2016colorful,vondrick2018tracking}, autoencoding and generating images~\cite{pathak2016context,donahue2016adversarial}.

Another form of self-supervised learning employs self-generated labels and trains on them. Our method falls in this latter category~\cite{bertinetto2016fully, dong2018triplet}. The most closely related work is that of Wang \textit{et al.}~\cite{wang2015unsupervised} which tracks objects in unlabeled videos and trains using a triplet loss and hard negative mining. However, this work considers visual tracking in 2D while we consider self-supervised 3D tracking. Additionally, it randomly samples from millions of class-agnostic patches for negative examples. However, we only use detections of the same object class obtained from an object detector; negative examples from such detections are likely to be harder and more useful for learning.

\subsection{3D Embeddings for Tracking}
Prior work has used neural network embeddings from point clouds for tracking. Zhang \textit{et al.}~\cite{zhang2019robust} uses a combination of 2D image embeddings and point cloud embeddings, via a min-cost flow graph technique. Giancola \textit{et al.}~\cite{giancola2019leveraging} use 3D point cloud embeddings in a single-object tracking setting.
Zarzar \textit{et al.}~\cite{zarzar2019efficient} builds on top of Giancola \textit{et al.}~\cite{giancola2019leveraging} using 2D birds-eye view for proposal generation and 3D PointNet based shape-completion encoder for matching.
However, in these approaches, training is fully supervised. In contrast, we train in a self-supervised manner using an uncertainty-aware loss, which enables our method to scale to large amounts of unlabeled training data.



\section{Approach}
\label{sec:approach}

\subsection{Overview}
We propose using self-supervision for 3D data association to leverage large amounts of unlabeled data. In this section, we describe our method, which consists of running a detector on unlabeled data, performing self-supervised data associations, and finally learning a point cloud embedding using triplet loss~\cite{chechik2010large}, such that embeddings are similar for detections of the same object in different views, but are dissimilar for detections of different objects.  We also describe how we use uncertainty-aware loss weighting and hard negative mining to obtain a more robust embedding.

\subsection{Obtaining Detections by Pre-trained Detector}
\label{sec:prob_mot}
We run a pre-trained 3D object detector on unlabeled data to obtain detections; in this work, we consider 3D tracking of cars in urban driving scenarios. We assume that we have an unlabeled dataset $U$ that consists of sequences of point clouds, such as those obtained from LIDAR sensors mounted on cars as they drive in urban environments. The pre-trained detector estimates the locations of 
3D bounding boxes around cars in each frame of each sequence in $U$. Although some of these detections may be incorrect, we treat these estimates as pseudo-ground truth labels for the purposes of self-supervision. 

\subsection{Obtaining Associations by Self-Supervision}
To obtain tracks of individual cars, we need to associate 3D bounding boxes of cars across multiple frames such that all associated boxes correspond to a unique car instance. We do not have access to ground truth associations in our unlabled dataset $U$. Instead, we adopt a Kalman Filter based method for multi-object association as proposed by Chiu \textit{et al.}~\cite{chiu2020probabilistic}. Each object’s state is modeled by a tuple of 11 variables: $s_t=(x,y,z,a,l,w,h,v_x,v_y,v_z,v_a)$, where $(x,y,z)$, represents the 3D object center position, $a$ represents the orientation of the object's bounding box, $(l,w,h)$ represent the length, width, and height of the object's bounding box, and $(v_x,v_y,v_z,v_a)$ represent the linear and angular velocity. The pre-trained object detector models each object's observation with a tuple of 7 variables: $o_t=(x,y,z,a,l,w,h)$. 

Using the Kalman filter, we can associate bounding boxes in consecutive frames based on the distance between each pair of predicted object state and observed object state. Specifically, we adopted Mahalanobis distance~(\cite{mahalanobis1936generalized,chiu2020probabilistic}) to measure the distance between predictions and detections, and employ a greedy algorithm with an upper bound threshold value to solve this problem. This is common practice in the tracking-by-detection approach of multi-object tracking~\cite{MOTChallenge2015}. Using the Mahalanobis distance relies on the assumption that objects in consecutive frames don't move very far, and prediction-detection pairs that have small Mahalanobis distance are likely to be of the same object. Although this assumption will sometimes be incorrect, we will show how we can use uncertainty to make our method more robust to errors in association.  


\subsection{Metric Learning via Triplet Loss}
Given a set of estimated detection and associations, we wish to learn a 3D embedding for an arbitrary point cloud inside a bounding box. We define two properties that the embedding must satisfy:
\begin{enumerate}
    \item Embeddings of bounding boxes of the same object across frames must be similar.
    \item Embeddings of bounding boxes of different objects across frames must be dissimilar.
\end{enumerate}
The similarity can be measured using cosine distance of the embedding i.e.
\newcommand{\eanc}{\mathbf{e}_\text{a}^t}
\newcommand{\epos}{\mathbf{e}_\text{p}^{t+k}}
\newcommand{\eneg}{\mathbf{e}_\text{n}^t}
\begin{equation}
\cos(\mathbf{e_1}, \mathbf{e_2}) = \frac{\mathbf{e_1} \cdot \mathbf{e_2}}{||\mathbf{e_1}|| \cdot ||\mathbf{e_2}||}
\end{equation}
where $\mathbf{e_1}, \mathbf{e_2}$ are 3D embeddings of two different point clouds.  Such  embeddings will be used to associate detected 3D bounding boxes at test time for 3D tracking.

In order to learn such an embedding, we use the triplet loss. This is defined as follows. Let $\eanc$ be the embedding of an `anchor' detection in frame $t$, and $\epos$ be the embedding of the same object in another frame $t+k$. Let $\eneg$ be the embedding of a different object in frame $t$ (why we choose the different object from the same frame $t$ will be discussed shortly). We shall refer to $(\eanc, \epos)$ as a positive pair and $(\eanc, \eneg)$ a negative pair. Then the triplet loss is defined as:
\begin{equation}
L(\eanc, \epos, \eneg) = \max \Big( \cos(\eanc, \eneg) - \cos(\eanc, \epos) + M, 0 \Big)  
\end{equation}
where $M$ is a margin hyperparameter (we found $M = 0.2$ to work well in practice).
The triplet loss is zero when the cosine similarity between the positive pair is at least $M$ more than the cosine similarity between the negative pair. The triplet loss encourages  embeddings of positive pairs to be more similar to each other than embeddings of negative pairs.  


An illustration of positive and negative pairs in the triplet loss of our self-supervised learning framework is shown in Figure \ref{fig:pull}. Given our estimated tracklets (i.e. detections and correspondences), we pick a tracklet at random and then we pick a pair of frames at random. Consider the frames at times $t$ and $t+k$. We use the detections of the tracked object in these frames as the positive pair in the triplet loss (shown in red in Figure \ref{fig:pull}). It is possible that this estimated association is incorrect and they do not correspond to the same object; we will describe how these errors can be overcome by incorporating an uncertainty-aware loss function.

For the negative pair, we pick any other detection from frame $t$ (shown in blue in Figure \ref{fig:pull}). Since this detection is in the same frame as the anchor detection (also from frame $t$), it is guaranteed to be a different object. We shall treat this as the negative pair.
Next, we crop and align point clouds inside the positive and negative pairs and train a 3D embedding using the triplet loss, as defined above.


\subsection{Incorporating Uncertainty}
\label{sec:uncertainty}

One of the most prominent challenges in training on labels generated via self-supervision is that the self-supervised data associations are not guaranteed to be accurate. 
Training on such erroneous labels may result in learning a 3D embedding that doesn't properly distinguish between the same object and different objects, reducing downstream tracking performance.

Our solution is to leverage \textit{uncertainty} in the self-supervised association labels. For examples with high uncertainty, we discount the contribution of the corresponding example during training. This is intended to reduce the effect of incorrect labels in self-supervised training. 

\subsubsection{Estimating Association Uncertainty}
\begin{figure}
    \centering
    \includegraphics[width=0.6\linewidth]{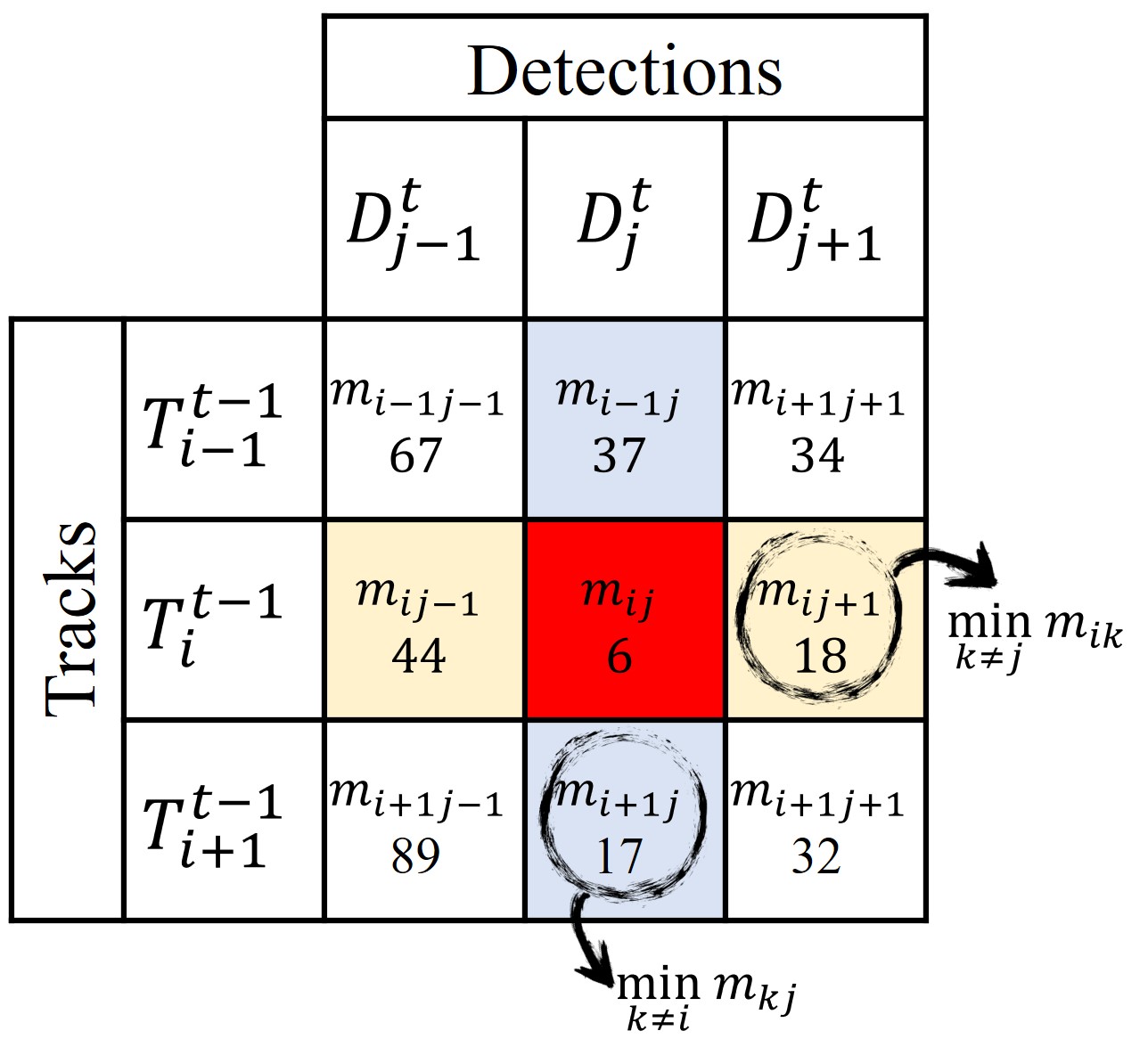}
    \caption{Illustration of estimating association uncertainty. We look at the Mahalanobis distances between all possible pairwise associations, greedily assign detections to tracks based on the lowest distance, and compute uncertainty by comparing with the lowest un-associated distance if we were to choose a different track or detection.}
    \label{fig:confidence_illustration}
\end{figure}

We propose a method for estimating the amount of uncertainty in the estimated data associations during training time. Given a set of previous tracks $\{T_i^{t-1}\}$  in frame $t-1$, and detections $\{D_i^t\}$ in frame $t$, we need to first associate each track with a detection. We predict the future state $\hat{\mu}_i^t$ for each track $T_i^{t-1}$ in frame $t$ using the Kalman filter, as well as the prediction uncertainty $\Sigma_i^t$. Then the Mahalanobis distance $m_{ij}$ is computed for every possible association pair $(T^{t-1}_i, D_j^t)$. The greedy algorithm defined in~\cite{chiu2020probabilistic} is used for performing data association. The algorithm is illustrated in Figure~\ref{fig:confidence_illustration} and proceeds as follows. First, the track $T_i^{t-1}$ that corresponds to the smallest Mahalanobis distance $m_{ij}$ is greedily associated with the respective detection $D_j^t$. Then, all pairs $m_{kl}$ with $k=i$ or $l=j$ are discarded, since $T_i^{t-1}$ and $D_j^t$ have already been associated to each other; we assume that each track can be associated to at most one detection, and vice versa. The greedy algorithm proceeds similarly until all tracks or all detections have been associated. Once a given association $(T^{t-1}_i, D_j^t)$ has been performed, we can define the confidence $AC$ of the associated pair  as following:
\begin{equation}
AC = 1 - \exp\Bigg(-\min\Big(\frac{\min_{k\neq j} m_{ik}}{m_{ij} + \epsilon}, \frac{\min_{k\neq i} m_{kj}}{m_{ij} + \epsilon}\Big)\Bigg)
\label{eq:ac}
\end{equation}
The quantity $m_{ij}$ denotes the Mahalanobis distance between the prediction of the associated track $T_i^{t-1}$ and detection $D_j^t$ (red cell in Figure~\ref{fig:confidence_illustration}). 
The ratio $\frac{\min_{k\neq j} m_{ik}}{m_{ij}}$
captures the confidence of associating track $T_i^{t-1}$ to detection $D^t_j$ as opposed to associating with another detection $D^t_k$. In other words, the quantity $\min_{k\neq j} m_{ik}$ denotes the smallest distance of associating the track $T_i^{t-1}$ with another candidate detection in frame $t$ that is not $D^t_j$ (yellow square with a circle in Figure~\ref{fig:confidence_illustration}). Thus, we are comparing the scores of associating track $i$ with detection $j$ compared with associating track $i$ with any other detection $k$.
The ratio $\frac{\min_{k\neq j} m_{ik}}{m_{ij}}$ will be large if the distance between $i$ and $j$ is much closer than the distance between $i$ and $k$. In such a case, one would be more confident that the association between $i$ and $j$ is correct, and our association confidence $AC$ will be close to 1.  On the other hand, if the distance between $i$ and $j$ is similar to or greater than the distance between $i$ and $k$, then the ratio $\frac{\min_{k\neq j} m_{ik}}{m_{ij}}$ will be low and the association confidence $AC$ will be smaller.


Similarly, the ratio $\frac{\min_{k\neq i} m_{kj}}{m_{ij}}$ captures the confidence of associating detection $D^t_j$ with track $T_i^{t-1}$ as opposed to associating with another track $T_k^{t-1}$. In order to be conservative, we consider the minimum of these two confidences to be the overall confidence. In order to normalize this value to lie in $[0, 1]$, we apply the monotonic operator $f(x) = 1-\exp(-x)$, that maps $[0, \infty)$ to $[0, 1]$. To avoid division by 0, we add $\epsilon=10^{-4}$ to $m_{ij}$ in practice .

The above described how we compute an association confidence between successive frames; we now describe how we compute such a confidence for more distant frames.
Consider predicted detections in frames  $t$ and $t+k$ that are arbitrarily far apart in time. Our confidence of these detections being correctly associated depends on how confident we are about the correctness of individual associations between successive frames $(t, t+1)$, ..., $(t+k-1, t+k)$. This is because if we are unsure about the association at any intermediate step, then our confidence of association across the $k$ frames should reduce. Hence, we define the cumulative association confidence (CAC) across frames $t$ and $t+k$ to be
\begin{equation}
CAC(t, t+k) = \Pi_{i=t}^{t+k} AC(t)
\label{eq:cac}
\end{equation}
where $AC(t)$ is the association confidence between frames $t$ and $t+1$, defined as in equation~\ref{eq:ac}.


\subsubsection{Uncertainty Weighted Triplet Loss}

Consider a triplet of embeddings ($\eanc, \epos, \eneg$) from frames $t$ and $t+k$. Since the negative pair $(\eanc, \eneg)$ comes from the same frame, we can be reasonably sure that they correspond to different objects. However, there is uncertainty in association of the positive pair $(\eanc, \epos)$ across $k$ frames. In particular, we estimated this certainty to be $CAC(t, t+k) \in [0, 1]$, as shown in Equation~\ref{eq:cac}. While training, we weight the triplet loss of each example by its association certainty:
\begin{equation}
L_{weighted}(\eanc, \epos, \eneg) = L(\eanc, \epos, \eneg) \times CAC(t, t+k)
\label{eq:weighted_loss}
\end{equation}
The total loss is given by the sum of losses from Equation~\ref{eq:weighted_loss} for each example.
Thus, for examples with a lower confidence of association, such examples will have a lower contribution to the total loss. Hence, the examples with highest certainty will end up contributing most to the loss. This is illustrated in Figure~\ref{fig:pipeline} (left) where we compute the confidence of a positive pair and use that to weight the triplet loss for that training example.


\subsection{Hard Negative Example Mining}
\label{section:hn}
Hard negative example mining~\cite{shrivastava2016training,felzenszwalb2009object} has shown to help improve accuracy and speed up training for object detection systems. In this technique, the hardest negative examples are first mined and then used to compute the optimization loss.

We use hard negative mining to select the negative example $\eneg$ in our triplet loss. Specifically, given an anchor embedding $\eanc$ in frame $t$, we compute the cosine similarity between $\eanc$ and the embedding $\mathbf{e}^t$ of every other detection in frame $t$. We choose the detection that has the highest cosine similarity with $\eanc$ since that is the hardest negative example:
\begin{equation}
    \mathbf{e}_n^t = \arg \max_{\mathbf{e}^t} \big\{\cos(\eanc, \mathbf{e}^t) \big\}
    \label{eq:hnm}
\end{equation}
This is illustrated in Figure~\ref{fig:pipeline} (left), in frame $t$. Multiple possible negative pairs in frame $t$ are shown;  the one whose embedding has the highest similarity with the anchor embedding is chosen for training. The algorithm is described in detail in Algorithm~\ref{algo:algo}. 

Note that our method does not require any annotated bounding box labels or association labels for the dataset $U$ on which the embedding is learned.  Instead, we use a pre-trained detector to predict bounding boxes and we use a Kalman filter to estimate associations.  Due to potentially erroneous associations, we use uncertainty weighting to downweight the loss of uncertain examples.  Thus, the embedding can be learned on a large unlabeled dataset; in this sense, we refer to our training procedure for the embedding as ``self-supervised," despite the fact that the detector itself was trained in a supervised manner.

\begin{algorithm}
\SetAlgoLined
 initialization: Embedding Function ($f_\theta$), \\Estimated Tracks $\{T\}_{i=1}^K$;\\
 \While{not converged}{
  sample a track $T \sim \{T\}_{i=1}^K$\\
  sample two detections $D_a^t, D_p^{t+k}\sim T$\\
  compute association uncertainty between $D_a^t, D_p^{t+k}$ using Eqn.~\ref{eq:cac}\\
  $D_n^t \leftarrow$ Hard Negative Mining ($D_a^t$) using Eqn.~\ref{eq:hnm}

  Train $f_\theta$ with loss defined by Eqn.~\ref{eq:weighted_loss}
 }
 \caption{Self-supervised 3D Data Association}
 \label{algo:algo}
\end{algorithm}

\section{Experiments}

In this section, we evaluate our approach on two tasks: single object tracking on the KITTI~\cite{geiger2013vision} dataset, and multi-object tracking on the NuScenes~\cite{nuscenes2019} dataset. We also verify the effectiveness of each component of our system by evaluating our proposed uncertainty estimation and performing ablation analysis.

\subsection{Self-supervised training of appearance embeddings}

\paragraph{Datasets}

For self-supervised training, we use the NuScenes~\cite{nuscenes2019} dataset. To obtain estimated detections, we evaluate the CBGS car detector~\cite{zhu2019class} on all frames in NuScenes. For evaluation, we use the KITTI~\cite{geiger2013vision} dataset in a single object tracking framework as described in the next section. 


\paragraph{Training Details}

During training, we generate triplets of training detections as our training instances, as described in Section~\ref{sec:approach}. The cropped point clouds are centered and aligned using the bounding box's location and pose. We use the PointNet~\cite{qi2017pointnet} architecture for extracting an embedding from a point cloud. We use the global feature of the PointNet as the learnable embedding, whose dimensionality is set to 1024. We use the Adam optimizer~\cite{kingma2014adam} for training the network with a batch size of 64 and a learning rate of $2\times10^{-5}$. We train the network for 100 epochs and report performance on the test set.

\subsection{Evaluation of Single Object Tracking}
\label{sec:evaluation}

\begin{table*}[h!]
    \centering
    \begin{tabular}{||c|c|c|c||} 
     \hline
        Method Type & Training Scheme & Accuracy without Noise (\%) & Accuracy with Noise (\%)\\
     \hline\hline
     Oracle (Fully supervised) & Ours (GT Detection + GT Association) & 42.9 & 39.4\\
     \hline
     \textbf{Ours (Weakly supervised)} & \makecell{Ours (GT Detection + SS Association)} & 42.7 & 38.9 \\
     \hline
     \textbf{Ours (Self supervised)} & \makecell{Ours (Est. Detection + SS Association)} & 42.4 & 39.5 \\
     \hline
     Baseline (Fully supervised) & ShapeCompletion3DTracking  & 42.3 & 36.8\\
     \hline
     Baseline & ShapeNet  & 33.7 & 32.0\\
     \hline
     Baseline & \makecell{Randomly Initialized Network} & 38.8 & 38.4\\
     \hline
     Baseline & Random classification & 25.3 & 25.3 \\
     \hline
    \end{tabular}
    \caption{Main Results. GT: ground truth, Est.: estimation, SS : self-supervised. Oracle performance is using ground truth detections and ground-truth associations during self-supervised training. Accuracy is shown for evaluation without and with noise added to candidate ground-truth detections during test time.}
    \label{table:main-results}
\end{table*}

\begin{table*}[h!]
    \centering
    \begin{tabular}{||c|| c| c|| c ||} 
     \hline
        Self-Sup. Training Scheme & Uncertainty & Hard Negative Mining & Accuracy (\%) \\ [0.5ex] 
     \hline\hline
     \multirow{4}{*}{\makecell{Weakly supervised: \\ GT Detection + SS Tracking}}
     & \xmark &  \xmark & 40.3 \\
     & \cmark &  \xmark & 41.9 \\
     & \xmark &  \cmark & 41.4 \\
     \cline{2-4}
      & \cmark &  \cmark & \bf{42.7} \\
     \hline\hline
     \multirow{4}{*}{\makecell{Self-supervised: \\Est. Detection + SS Tracking}} 
     & \xmark &  \xmark & 39.7 \\
     & \cmark &  \xmark & 41.4 \\
     & \xmark &  \cmark & 41.3 \\
     \cline{2-4}
    & \cmark &  \cmark & \bf{42.4} \\
     \hline
    \end{tabular}
    \caption{Ablation experiments removing one component of our method at a time. Table shows that incorporating uncertainty, hard negative example mining, gives the best performance in both the self-supervised training schemes.}
    \label{table:ablations}
    \end{table*}

We evaluate the learned self-supervised 3D embeddings in a single-object tracking framework, similar to~\cite{giancola2019leveraging,zarzar2019efficient}, where only one object is tracked at a time. 

We adapt the training set of the KITTI tracking dataset~\cite{geiger2013vision} for single object tracking, similar to~\cite{giancola2019leveraging,zarzar2019efficient}. We use the same test set as~\cite{giancola2019leveraging} i.e. scenes 19-20 in KITTI. Tracklets are generated for each instance of a car appearing in each scene. Each tracklet consists of frames in a scene in which a given car appears. Only the first frame is provided with the ground truth 3D bounding box.

We wish to evaluate the quality of the learned 3D embeddings for the association task. Thus, for our evaluation, we assume that ground truth detections of all cars in the scene are provided. This allows us to avoid confounding errors between detection and data association and allows us to evaluate the improvements in data association in an isolated manner.


However, we additionally wish to evaluate how robust the learned embedding is to noisy detections. Thus, we perform a separate evaluation in which we add random noise to the location, size and orientation of the ground-truth detections. The location and size noise are uniformly sample from $\pm 10\%$ of the bounding box size, while orientation noise is uniformly sample from $\pm 5^{\circ}$.


As is typical in single-object tracking, we assume that the ground-truth bounding box  of the object to be tracked is provided in the first frame of the point cloud sequence. We use this bounding box as the  anchor to be matched to in subsequent frames. In the $t$-th frame, the embedding for each of the candidate ground truth bounding boxes is computed. The candidate whose embedding has the highest cosine similarity score with the anchor embedding is selected. This process is illustrated in Figure~\ref{fig:pipeline} (right). Note that this is essentially a classification task: exactly one of the candidate bounding boxes corresponds to the correct car being tracked. We thus evaluate the embedding using the classification accuracy of associating the correct bounding box. A 3D embedding that is able to discriminate between the same object in future frames versus different objects will produce a high classification score.



\paragraph{Results}

Table \ref{table:main-results} shows the classification accuracy of association for various approaches to learning an embedding. We evaluate two different versions of our method:
\begin{itemize}
    \item Ours (Weakly supervised): First, we assume that ground truth detections are available during training (GT Detection), but association labels are not available and must be self-supervised  (SS Association) using our method, along with uncertainty-weighting and hard negative mining. This corresponds to row 2 of Table~\ref{table:main-results}.
    \item Ours (Self supervised): Second, we assume that neither ground truth detections not ground truth associations are available on the dataset on which the embedding is learned. In this setting, detections must be estimated using pre-trained detector (Est. Detection) and associations are self-supervised (SS Association) using our method with uncertainty-weighting and hard negative mining. This corresponds to row 3 in Table \ref{table:main-results}.  
\end{itemize}



We compare our results to the following baselines:
\begin{itemize}
    \item \textit{ShapeNet}: This is the global feature of PointNet trained to classify shapes in ShapeNet~\cite{qi2017pointnet}, including cars. 
    \item \textit{ShapeCompletion3DTracking~\cite{giancola2019leveraging}}: This embedding was trained on KITTI~\cite{geiger2013vision} in a fully supervised manner with ground-truth detections and ground-truth associations using a combination of reconstruction loss and cosine similarity loss. 
    \item \textit{Randomly Initialized Network Embedding}: As a sanity check, we randomly initialize the weights of the PointNet architecture. 
    \item \textit{Random Classification}: As another sanity check, we randomly pick one of the candidate ground-truth detection in frame $t$. An accuracy of about 25\% suggests that there are four candidates on an average per frame.
\end{itemize}
As shown in Table~\ref{table:main-results}, our method, trained on unlabeled data, outperforms all of these baselines and is especially robust to detection noise.  The method of ShapeCompletion3DTracking~\cite{giancola2019leveraging} has performance near to that of our method; however their approach was trained on fully supervised data with ground-truth detections and ground-truth data associations.  On the other hand, our method (row 3) was trained on unlabeled data with estimated detections and estimated data associations, using uncertainty-weighting and hard negative mining.  We will show below that these are crucial components for our method.  

It should be noted that ShapeCompletion3DTracking~\cite{giancola2019leveraging} was trained on KITTI~\cite{geiger2013vision}, whereas our method was trained on NuScenes~\cite{nuscenes2019} so these results are illustrative but not directly comparable.  Regardless, the purpose of our work is to demonstrate that we can successfully train an embedding without requiring bounding box or association labels, which we have demonstrated.


To provide an upper bound on our performance, we also compare against using full supervision, i.e. training with ground-truth detections and ground-truth associations available in the NuScenes~\cite{nuscenes2019} dataset. This is the ideal case for self-supervised learning and achieves the oracle performance of our method. As shown in the table, the performance of our proposed method is very close to this oracle. However, because our method can be used to train an embedding on an unlabeled dataset, we could potentially obtain even better performance than this Oracle given a larger unlabeled dataset on which to train our method.


\paragraph{Ablations}
We perform an ablation analysis to quantify the performance of each component of our system. The results are shown in Table~\ref{table:ablations}. We perform ablations under each training regime: (1) Weakly supervised, using ground truth detections + self-supervised tracking, and (2) Self supervised, using estimated detections + self-supervised tracking. In both cases, we remove one component of our training procedure at a time. During evaluation, we do not add any noise to candidate ground truth bounding boxes. We note that either not incorporating uncertainty, or removing hard negative mining, consistently reduces performance. This shows that both these components are useful for our method.

\paragraph{Qualitative Analysis}
\begin{figure*}
    (a)
    \begin{subfigure}{0.60\textwidth}
    \newcommand{\width}{.3\textwidth}
    \makeatletter
    \define@key{Gin}{mycrops}[]{\setkeys{Gin}{trim=150 90 150 250,clip}}
    \makeatother
    \centering
    
    \hspace{-0.6cm} \small{Tracked Object} \hspace{1.8cm} Ours\hspace{2.3cm} ShapeNet
    
    \includegraphics[width=\width,mycrops]{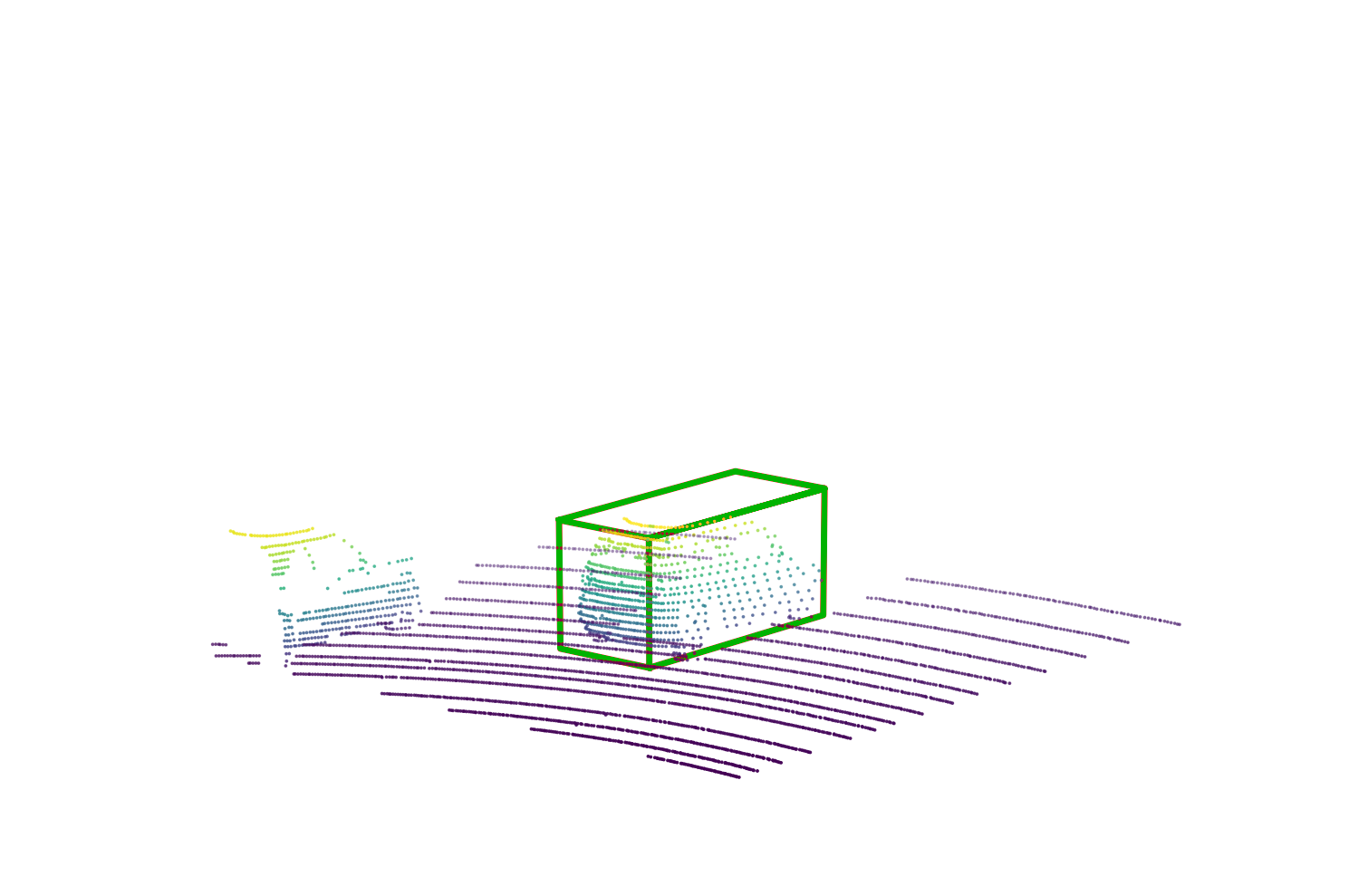}
    \includegraphics[width=\width,mycrops]{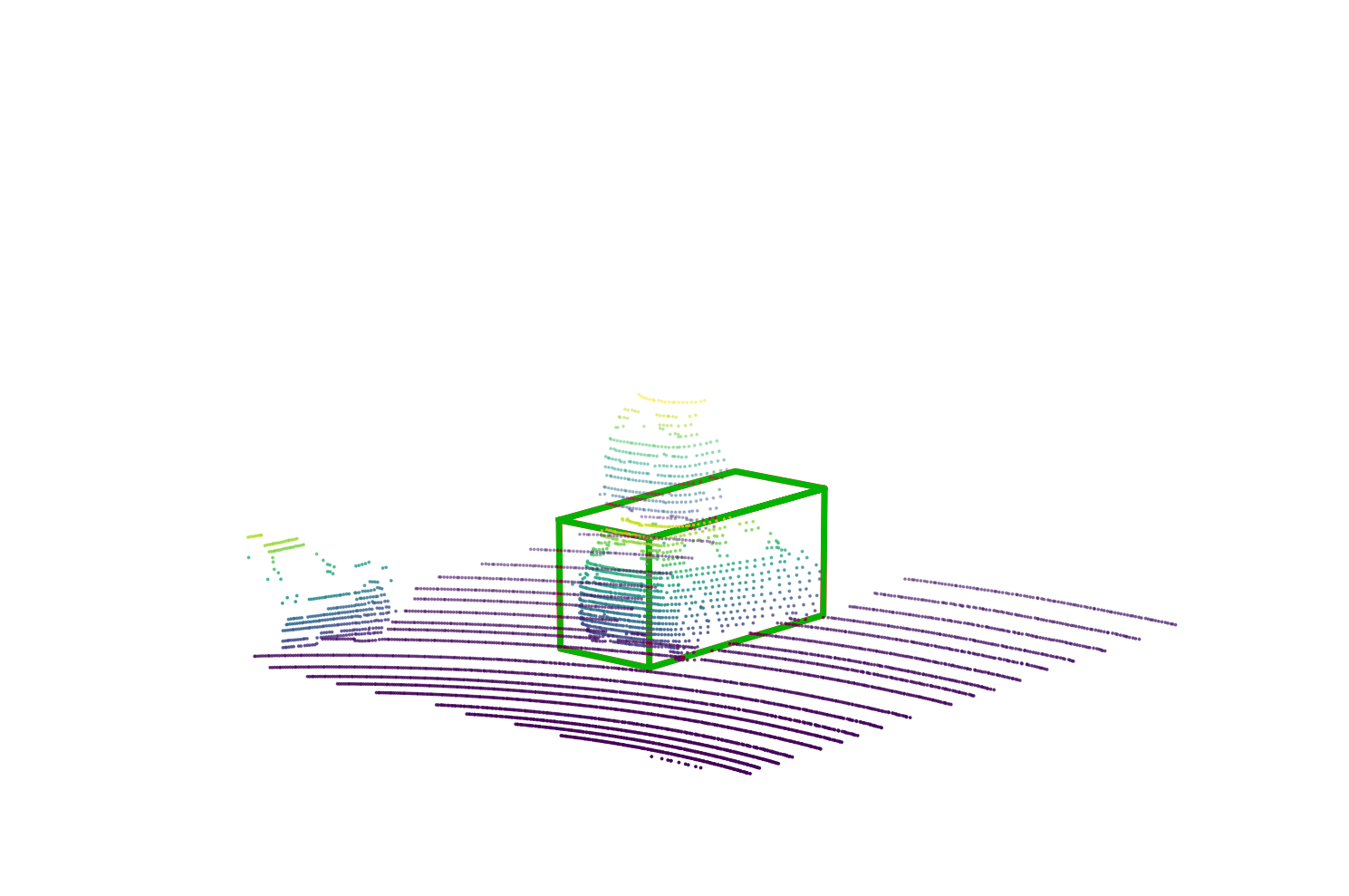}
    \includegraphics[width=\width,mycrops]{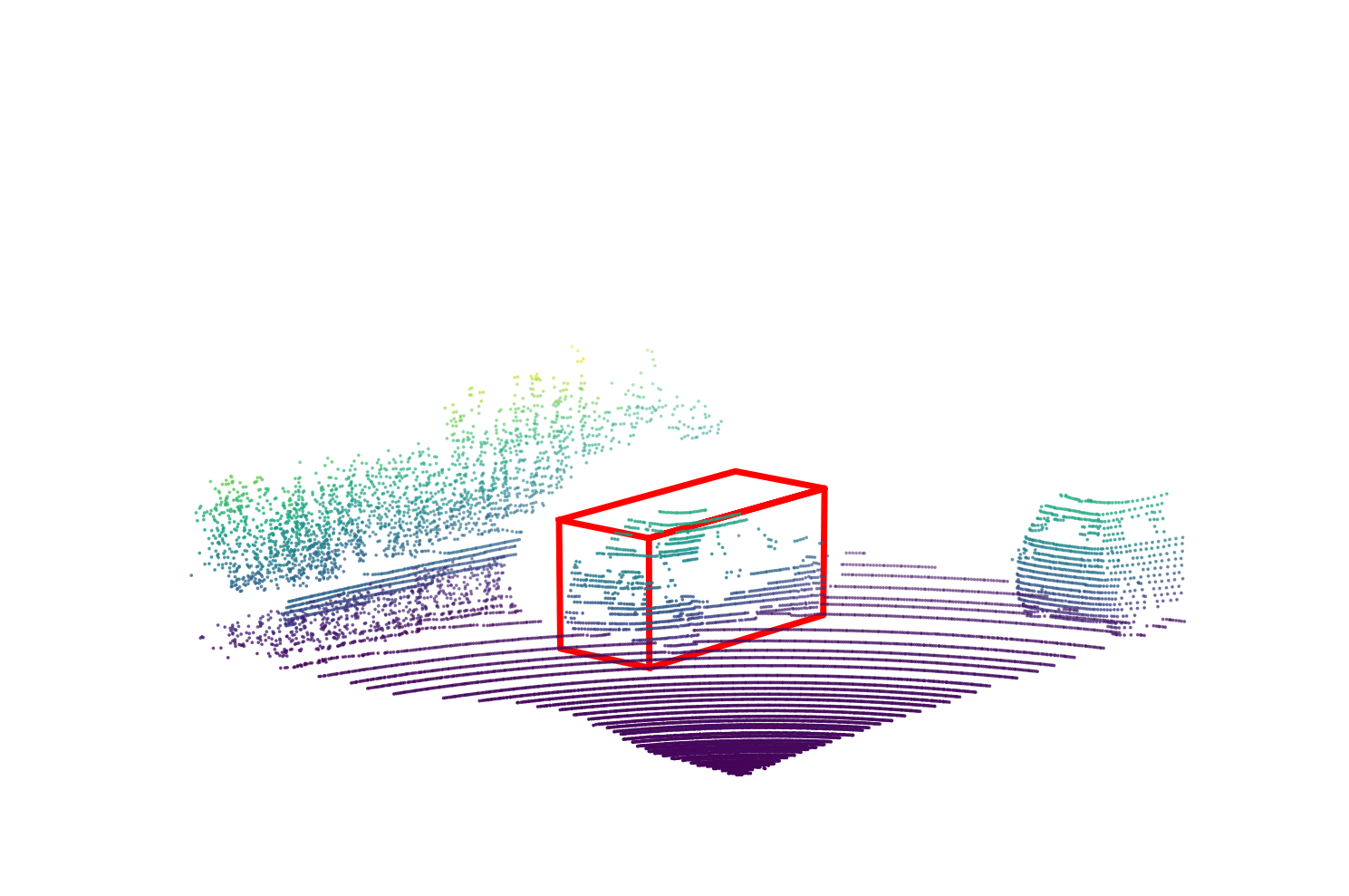}

    \includegraphics[width=\width,mycrops]{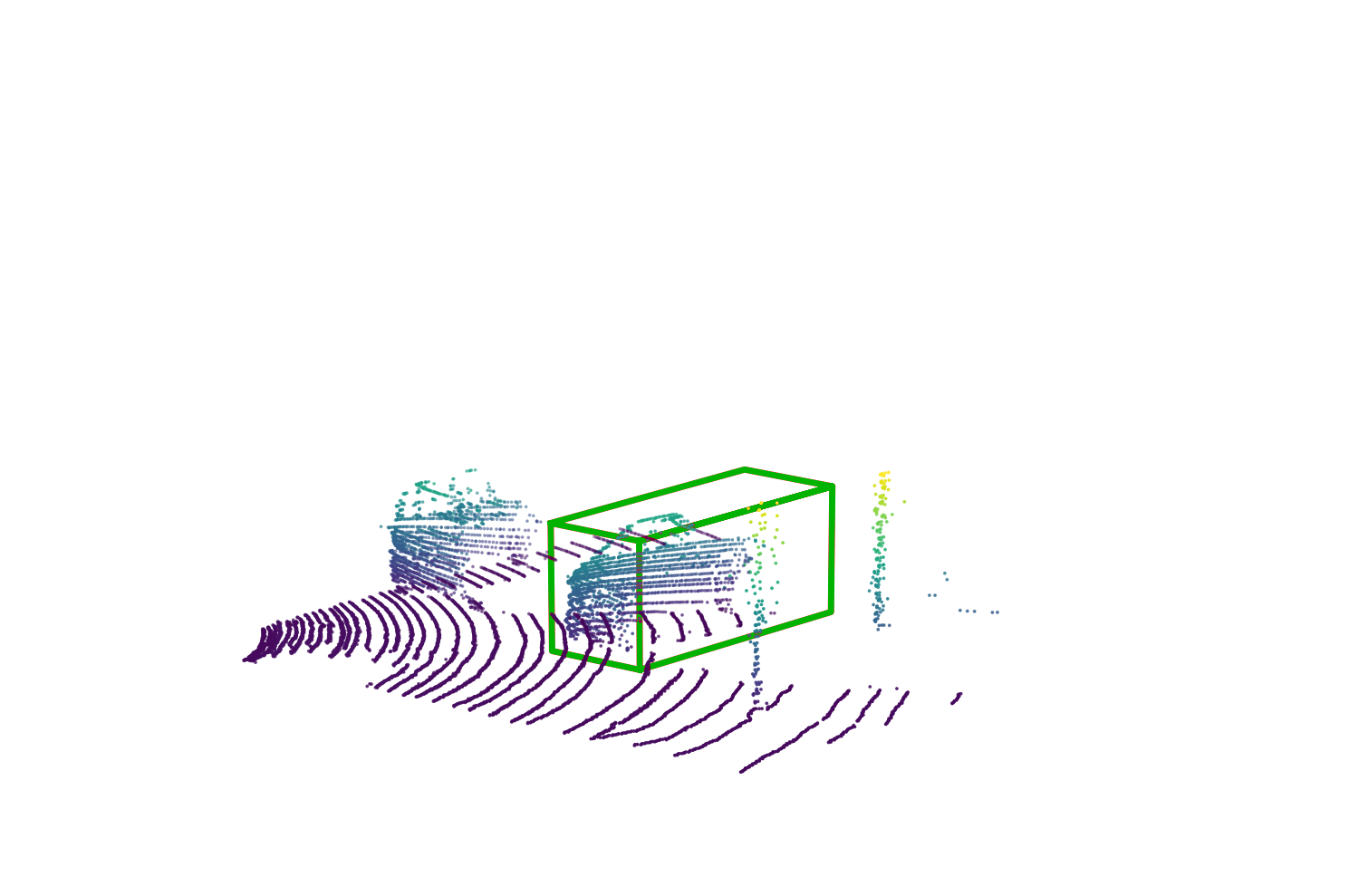}
    \includegraphics[width=\width,mycrops]{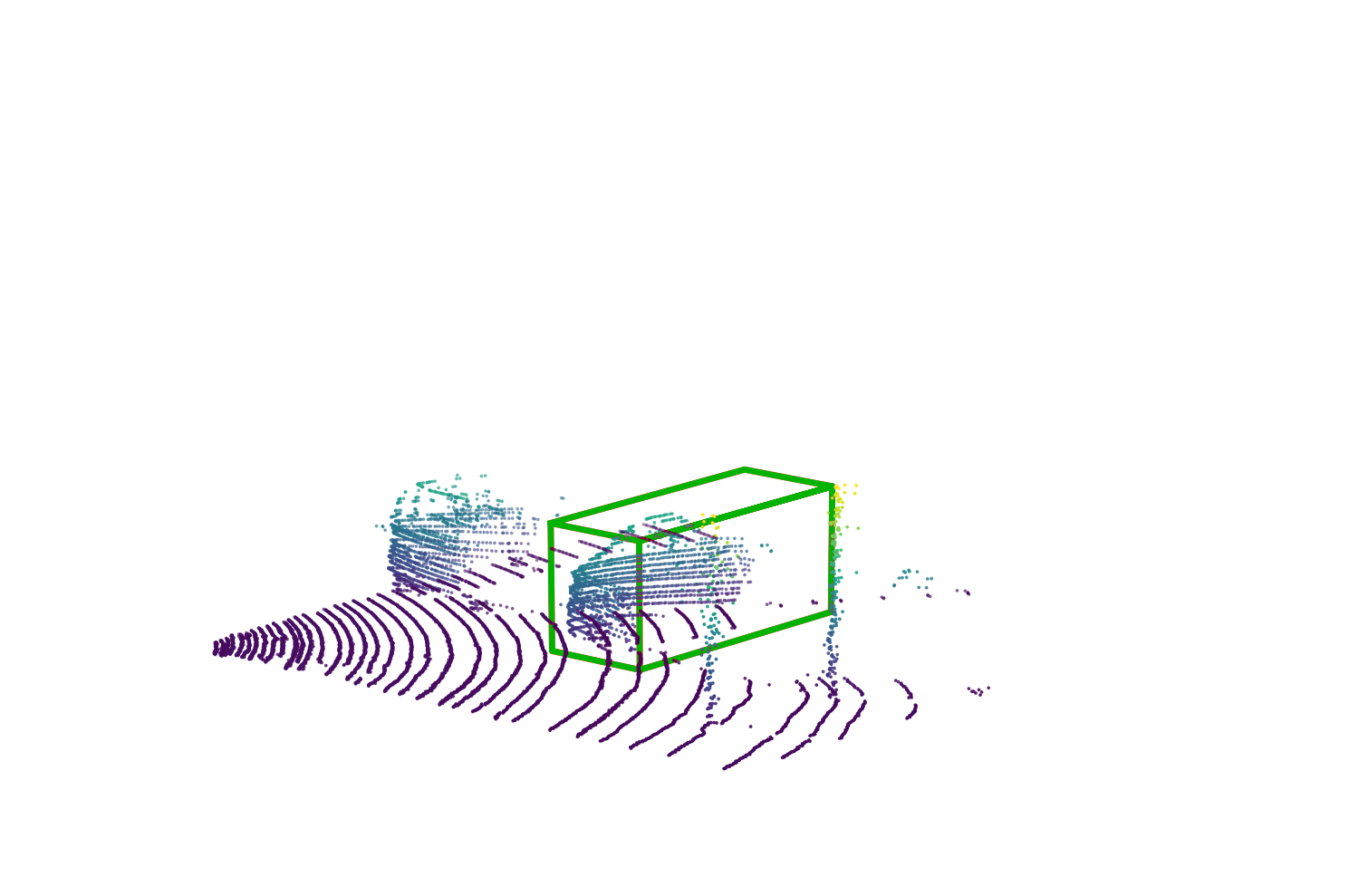}
    \includegraphics[width=\width,mycrops]{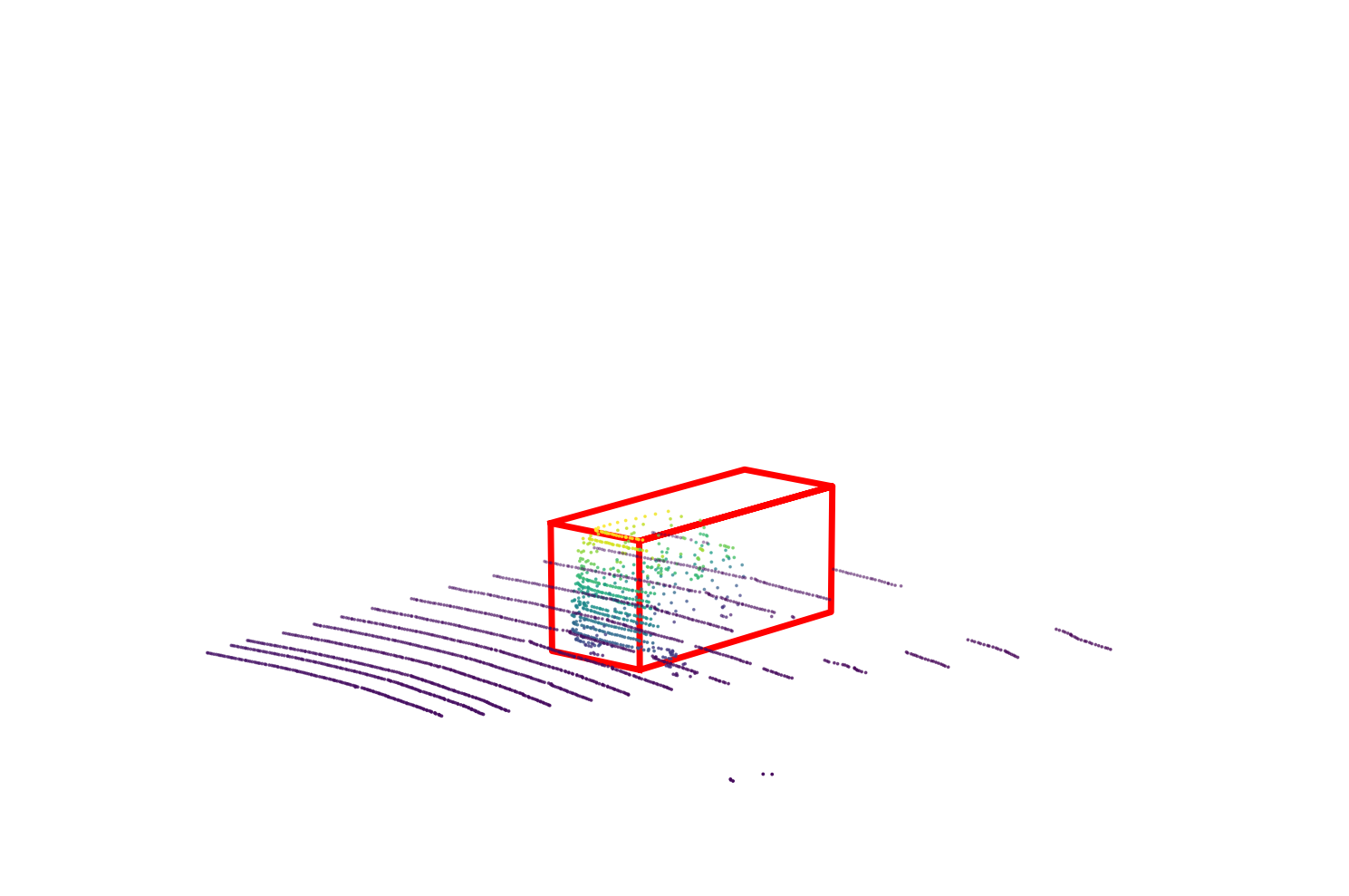}
  \end{subfigure}
  (b)
  \begin{subfigure}{0.34\textwidth}
    \includegraphics[width=\textwidth]{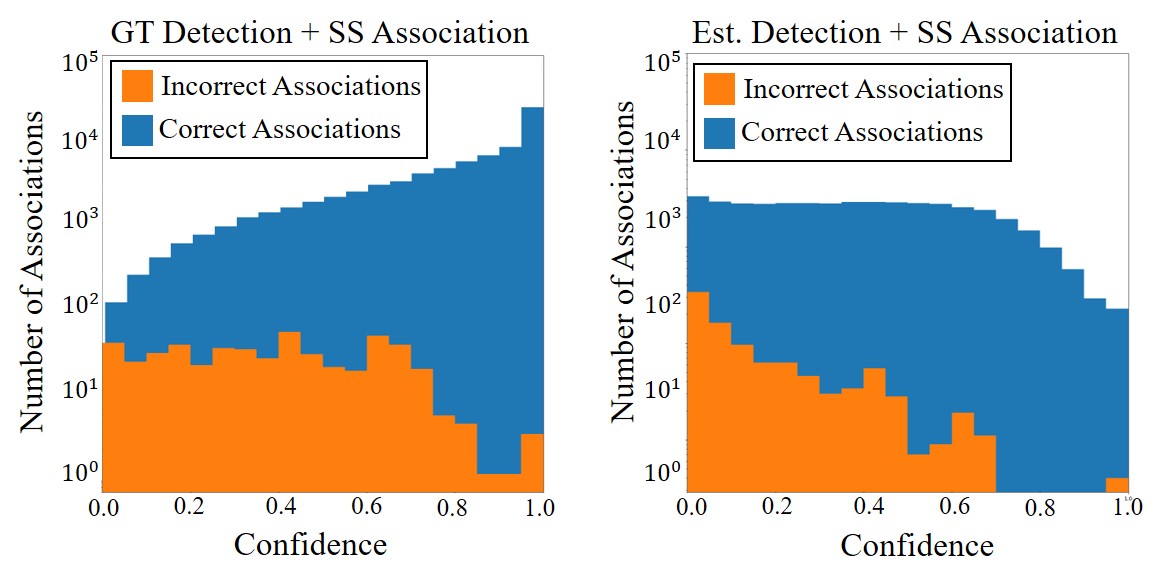}
  \end{subfigure}

    \caption{(a): \textit{Qualitative Analysis}. \textit{Left:} First frame of the tracklet, which defines the object to be tracked, \textit{Middle:} Bounding box selected by our embedding, trained using self-supervised detections and self-supervised tracking, while incorporating uncertainty and hard negative mining, \textit{Right:} bounding box selected by the baseline ShapeNet embedding. Boxes in the middle and right column are candidates from the same frame. In these examples, our embedding is able to select the correct bounding box, whereas the ShapeNet embedding selects another distractor object. (b): \textit{Evaluating uncertainty estimation}. Frequency of positive and negative association pairs as a function confidence. Correct association are correlated with high confidence and incorrect detections are correlated with low confidence.}

    \label{fig:qualitative}
\end{figure*}

In Figure \ref{fig:qualitative}(a), we show examples where the embedding obtained by estimated detection and self-supervised association selects the correct detection box in frame $t$, whereas the baseline ShapeNet embedding fails. The first column shows the exemplar anchor box. This denotes the car that is to be tracked. The middle column shows the bounding box of the same object in frame $t$ that is selected by our method. The right column shows bounding boxes of distractor objects in frame $t$ that the ShapeNet embedding deems to be more similar to the tracked object. These examples show that the embedding learned using a triplet loss via self-supervision is able to correctly match point clouds of the same object even if the distractor object's point cloud is only subtly different (Figure \ref{fig:qualitative}(a) rows 1 and 2).

\subsection{Evaluation of Multi-object Tracking}

\paragraph{Combining 3D appearance with motion and detection scores} We use the learned 3D embedding to improve multi-object tracking (MOT), which aims at performing data associating  for multiple targets.  The dominant strategy to this problem, \textit{i.e., tracking-by-detection} breaks MOT into two steps: 1) the detection step, where detections are first obtained by running an object detector on each frame (see Sec.~\ref{sec:prob_mot}). These detections are inputs to an MOT algorithm. 2) the association step, where detections in new frames are assigned to the existing trajectories. The trajectories may be created and terminated during the tracking process, which are the outputs of the MOT association algorithm.
For example, Chiu et. al.~\cite{chiu2020probabilistic} use the Mahalobis distance from a Kalman filter as a basis for association; the detection pairs that have a smaller Mahalanobis distance get associated, based on a greedy algorithm that achieves state-of-the-art tracking performance. We propose to improve the association by combining the Mahalanobis distance with our self-supervised 3D embedding. We also show benefit by integrating the detection scores into the association metric. 
For a track $T_i$ that has been tracked until frame $t$, the Kalman filter forecasts a bounding box $\hat{D}^{t+1}_i$ in the next frame ($t+1$).
Then, the overall association score $A_{ij}$ between $\hat{D}^{t+1}_i$ and a candidate detection $D^{t+1}_j$ in the next frame is expressed as a linear combination of $m, a, d, \log m, \log a, \log d$, where $m$ is the Mahalanobis distance between $\hat{D}^{t+1}_i$ and $D^{t+1}_j$, $a$ is the cosine between the 3D embeddings of $\hat{D}^{t+1}_i$ and $D^{t+1}_j$, and $d$ is the detection score of and $D^{t+1}_j$ .
The coefficients of the linear function are estimated via logistic regression which classifies whether $D^{t+1}_j$ belongs to the track $T_i$.
The logistic regression only needs a very small supervised dataset since the dimensionality of the feature space (only 6 features) is relatively small.
At test time, we use the greedy algorithm for association as specified in~\cite{chiu2020probabilistic}, where we use the logistic output $A_{ij}$ instead of only using the Mahalanobis score as in past work.

\paragraph{Dataset and setting}

We adopt the PanoNet3D~\cite{Chen-2020-121397} detection results and train a separate 3D embedding for each category using 90 scenes from the NuScenes~\cite{nuscenes2019} validation set. We estimate the logistic function parameters using 30 scenes from the NuScenes~\cite{nuscenes2019} validation set. We evaluate the performance of multi-object tracking on the remaining 30 scenes from the NuScenes~\cite{nuscenes2019} validation set. We follow the evaluation procedure of the NuScenes Tracking Challenge~\cite{nuscenes2019} and report the Average Multi-Object Tracking Accuracy (AMOTA)~\cite{Weng2020_AB3DMOT} as the main evaluation metric. 

\paragraph{Results}

The results for multi-object tracking are shown in Table~\ref{table:mot-results}. We use the following notation:
\begin{itemize}
    \item Motion: using only the Mahalanobis distance in the association score.
    \item Motion + Detection Score: using the Mahalanobis distance and the detection confidence as the association score.
    \item Motion + Detection Score + Appearance: using the Mahalanobis distance, detection confidence, and appearance similarity based on our self-supervised 3D embeddings as the association score.
\end{itemize}

Table~\ref{table:mot-results} shows that using our self-supervised 3D embeddings consistently improves the performance of multi-object tracking, across almost all object classes.


\begin{table*}[h!]
    \centering
    \begin{tabular}{||c|c|c|c|c|c|c|c|c||}
    \hline
    Method & Bicycle & Bus & Car & Motorcycle & Pedestrian & Trailer & Truck & Overall\\
    \hline 
     Motion & 30.3 & 72.8 & 74.5 & 56.4 & 75.6 & 45.3 & 54.2 & 58.4\\
     \hline
     Motion + Detection Score & 29.3 & 74.8 & 74.3 & 56.5 & 75.9 & \textbf{45.9} & 54.7 & 58.8\\
     \hline
     Motion + Detection Score + Appearance & \textbf{32.4} & \textbf{75.0} & \textbf{75.0} & \textbf{57.1} & \textbf{76.4} & 45.3 & \textbf{54.9} & \textbf{59.4}\\
     \hline
    \end{tabular}
    \caption{AMOTA scores for Multi-object tracking based on Probabilistic Tracking~\cite{chiu2020probabilistic} with different association scores. 
    }
    \label{table:mot-results}
\end{table*}

\subsection{Evaluation of Uncertainty Estimation}


We also verify the effectiveness of our proposed method for estimating association uncertainty. A useful association uncertainty metric is one that is uncertain for incorrect association pairs (association pairs with different ground truth identity), but is confident for correct association pairs (association pairs with same ground truth identity). We evaluate our uncertainty by observing how correlated it is with the correctness of associations. Figure~\ref{fig:qualitative}(b) shows histograms of the frequencies of association pairs in the nuScenes validation set, with respect to their estimated confidence. We can see that correct association are correlated with high confidence (low uncertainty) and wrong associations are correlated with low confidence (high uncertainty). In the ground truth detection and self-supervised association regime, the mean confidence for correct associations is 0.80, whereas it is only 0.39 for incorrect associations (Figure~\ref{fig:qualitative} (b, left)). In the estimated detection and self-supervised association regime, the mean confidence for correct associations is 0.40, whereas it is only 0.16 for incorrect associations (Figure~\ref{fig:qualitative} (b, right)). The correlation between our estimated uncertainty and the correctness of association indicates that it is useful for suppressing the contribution of incorrect associations during self-supervised training.

\section{Conclusion}

In conclusion, we propose a self-supervised representation learning method for 3D tracking. We show that by incorporating tracking uncertainty and hard negative mining, our learned point-cloud embeddings are effective for 3D association and tracking without using any human annotation, for both single-object and multi-object tracking. We hope that our work points towards moving away from spending excessive efforts annotating labeled data and instead redirecting them to self-supervised learning on large unlabeled datasets.

\section{Acknowledgements} This material is based upon work supported by the National Science Foundation under Grant No.  IIS-1849154, by the United States Air Force and DARPA under Contract No. FA8750-18-C-0092, and by the Honda Research Institute USA.

{\small
\bibliographystyle{IEEEtran}
\bibliography{egbib}
}

\end{document}